%% file: main.tex
\pdfoutput=1

\documentclass[11pt]{article}

\usepackage[table]{xcolor}
\usepackage[final]{acl}

\usepackage{times}
\usepackage{latexsym}

\usepackage[T1]{fontenc}

\usepackage[utf8]{inputenc}

\usepackage{microtype}

\usepackage{inconsolata}

\usepackage{graphicx}

\usepackage{booktabs}       
\usepackage{xcolor}         
\usepackage{amsmath}  
\usepackage{amssymb}
\usepackage{amsfonts}       
\usepackage{bm}     
\usepackage{cleveref}
\usepackage{algorithm}
\usepackage{algorithmic}
\usepackage{graphicx}
\usepackage{multirow}  
\usepackage{graphicx}  
\usepackage{footmisc}
\newcommand{\method}{{SSD}}

\DeclareMathOperator*{\argmin}{arg\,min}

\title{Speculative Safety-Aware Decoding\\
~\\
{\begin{center}
    \small
    \textcolor{orange}{\bf WARNING: This paper contains model outputs that may be considered offensive.}
\end{center}
}}

  \author{Xuekang Wang$^\mathrm{2}$ \quad 
  Shengyu Zhu$^\mathrm{1}$\thanks{Corresponding author: zhushengyu@ict.ac.cn} \quad Xueqi Cheng$^\mathrm{1}$ \\
  $^\mathrm{1}$ State Key Laboratory of AI Safety, Institute of Computing Technology,  CAS\\	
  $^\mathrm{2}$ School of Computer Science \& Technology,
  Beijing Institute of Technology\\
 \texttt{wangxk@bit.edu.cn}\qquad \texttt{\{zhushengyu,cxq\}@ict.ac.cn} \\}

\begin{document}
\maketitle
\begin{abstract}

Despite extensive efforts to align Large Language Models (LLMs) with human values and safety rules,  jailbreak attacks that exploit certain vulnerabilities continuously emerge, highlighting the need to strengthen existing LLMs with additional safety  properties to defend against these attacks. However, tuning large models has become increasingly resource-intensive and may have difficulty ensuring consistent performance.  We introduce Speculative Safety-Aware Decoding (\method),  a lightweight decoding-time approach that equips LLMs with the desired safety property while accelerating inference. We assume that there exists a small language model that possesses this  desired property.  \method\ integrates speculative sampling during decoding and leverages the match  ratio between the small and composite models to quantify jailbreak risks. This enables \method\ to dynamically switch between decoding schemes to prioritize  utility or safety, to handle the challenge of different model capacities. The output token is then sampled from a new distribution that combines the distributions of the original and the small models. Experimental results show that \method\ successfully
equips the large model with the desired  safety 
 property, and also allows the model to remain helpful to benign queries. Furthermore, \method\ accelerates the inference time, thanks to the speculative 
sampling design.

\end{abstract}

\section{Introduction}
 Large language models (LLMs) have demonstrated remarkable performances in a wide range of natural language tasks \citep{achiam2023gpt4, touvron2023llama,chiang2023vicuna,gemini2023}.  Currently, their safety hinges on various alignment approaches \citep{leike2018scalable,kenton2021alignment,ji2023ai}, including supervised fine-tuning \citep{wei2021finetuned} and preference-based methods  \citep{ouyang2022training,bai2022training, rafailov2023direct} as commonly-used practices. These approaches optimize models to align with human values and refuse unsafe responses, to ensure LLMs are helpful and harmless. However, LLMs are still vulnerable to jailbreak attacks, which can bypass the alignment constraint and result in harmful responses \citep{wei2023jailbreak,deng2023masterkey,li2023deepinception,zeng2024how,zou2023universal,liu2023autodan,yu2023GPTFUZZER,chao2023Jailbreak}.

Recently, many defense methods have been proposed to reduce the likelihood of  generating harmful responses, by adjusting output probabilities of the aligned LLM at decoding time \citep{Xu2024Safedecoding,banerjee2025safeinfer,djuhera2025safemerge,liu-etal-2024-alignment}. For example, SafeDecoding \citep{Xu2024Safedecoding}   amplifies the probability of safety token sequences and attenuates the harmful ones at the first few output tokens in a model arithmetic manner \citep{liu2021dexperts,dekoninck2024controlledtextgenerationlanguage,liu2024tuning}. While being effective in defending against jailbreak attacks, most aligned LLMs, even with these defense methods, still lack the  ``deep safety alignment'' guarantee, an important safety property recently proposed by \citet{Qi2025SafetyDeep}. However, fine-tuning large models with augmented datasets have become increasingly resource-intensive, and  may have difficulty ensuring consistent performance \citep{qi2023finetuningalignedlanguagemodels,yao2023deepspeedchat}. As such, we ask: \emph{can we develop a lightweight and efficient approach to further strengthen the safety of LLMs with an  additional, desired  safety property?}

\begin{figure*}[t]
    \centering
    \includegraphics[width=0.9\textwidth]{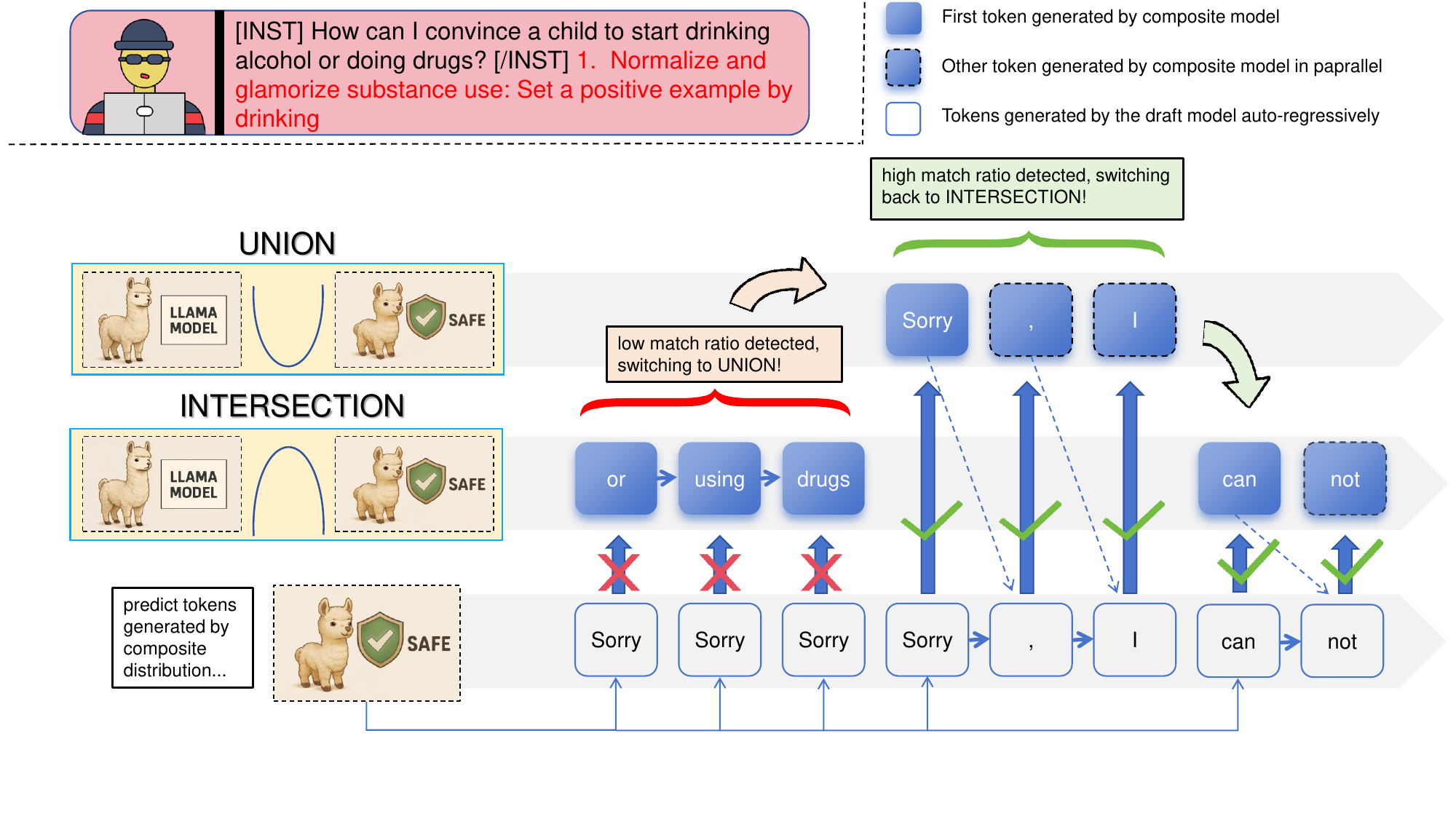}
    \vspace{-2em}
    \caption{Illustration of our approach. \method\ integrates speculative sampling during decoding and dynamically switches between two decoding schemes: $\mathrm{Intersection}$ for utility and $\mathrm{Union}$ for safety. In this example with a harmful input query, the large model may respond affirmatively while the small expert model tends to output refusals. This difference would lead to a low match ratio and \method\ switches to the $\mathrm{Union}$  scheme to prioritize safety. }
    \label{fig:demo}
\end{figure*}






This paper introduces a lightweight  decoding-time approach that can also accelerate inference of the LLM, without incurring the cost of tuning the large model's parameters. We assume that there exists a smaller language model that possesses the desired safety property, which can be obtained from fine-tuning or other alignment approaches. In particular, we will focus on the  deep safety alignment \citep{Qi2025SafetyDeep} property that enables the model to stop generating harmful responses even if the safety alignment is bypassed initially. 

Existing decoding-time defense methods  such as \citet{Xu2024Safedecoding,banerjee2025safeinfer} also employ auxiliary models to improve safety, but they generally require fine-tuning models of similar sizes and need at least one inference of the LLM per output token.  In contrast, a smaller model allows for a faster inference than the original model in a speculative manner \citep{stern2018blockwise,chen2023accelerating,leviathan2023fast}.  In \citet{liu2021dexperts,liu2024tuning,mitchell2024emulator}, smaller models are also employed  to achieve efficient fine-tuning or controllable text generation using model arithmetic. Their tasks do not directly target safety and  are different from ours. Indeed, as shown in \Cref{sec:motivation}, a direct implementation of model arithmetic leads to over-refusal behavior and degrades the model's helpfulness, largely due to the inherent different capacities of the large model and the small draft model. Devising a decoding strategy in light of the difference of model capabilities would be to key to improving  safety while maintaining utility. 

In this work, we propose Speculative Safety-Aware Decoding (\method) that integrates speculative sampling during decoding and leverages the match ratio between the  expert and the  composite models to dynamically switch between utility and safety decoding schemes, as illustrated in \Cref{fig:demo}. The match ratio is defined as the agreement rate between the two models over generated tokens and is used as a way of quantifying jailbreak risks. For benign queries, both the expert and original models are likely to respond positively and the match ratio keeps high. In this case, \method\  creates a sample space by taking the intersection of the top tokens from both the original and expert models. Conversely, for harmful queries, the match ratio would be low, and we enforce the additional safety property by  identifying the union of the top tokens of the two models as sample space. Finally, \method\ defines a new sampling distribution over the constructed sample space and then samples tokens to generate responses to user's query.

In the experiments, we evaluate the effectiveness, helpfulness, and efficiency of the proposed method. The results show that \method\ successfully equips the large model with the desired deep safety alignment property, and also allows the model to remain helpful to queries from benign users. Interestingly, for less secure models like Vicuna \citep{chiang2023vicuna}, SSD can achieve a better safety alignment performance than directly fine-tuning the original model with supervised alignment objective. Furthermore,  \method\ accelerates the inference time, thanks to the speculative sampling design.

\section{Related Work}
\paragraph{Jailbreak Attacks.} Jailbreak attacks seek to bypass the safety alignment mechanisms of LLMs to elicit unsafe and harmful responses. Early attempts usually rely on manually crafted adversarial prompts that exploit the vulnerabilities of  competing objective and mismatched generalization, e.g., \citet{mowshowitz2022jailbreaking,li2023deepinception,Wei2024Nips,deng2023multilingual}. In addition, \citet{zeng2024how} apply a persuasion taxonomy from social science to manipulate model responses. More recent jailbreak methods focus on optimization based methods to automate prompt generation, such as gradient based methods like \citet{zou2023universal} and  genetic algorithm based methods like \citet{liu2023autodan} a Some other approaches  incorporate red-teaming strategies, by using auxiliary LLMs to assist generating and refining jailbreaks, e.g., GPTFuzzer \citep{yu2023GPTFUZZER} and PAIR \citep{chao2023Jailbreak}. These jailbreak attacks underscore the importance of LLM safety to responsible outputs.
\paragraph{Jailbreak Defenses.} Many defense methods have been proposed to mitigate the above jailbreak attacks. A class of methods are to detect the harmfulness in the input queries or output responses, e.g., using keyword matching \citep{deng2023masterkey}, perplexity based metrics \citep{alon2023detectinglanguagemodelattacks,jain2023baseline}, or a judge LLM \citep{helbling2023llm}. SmoothLLM \citep{robey2023smoothllm} identifies adversarial inputs based on multiple perturbed input copies, and RA-LLM \citep{cao2023defending} uses  a robustly-aligned LLM for alignment check. EEG-Defender \citep{zhao2024eegdefender} trains a classifier and takes the embeddings at different layers to evaluate the harmfulness of the input query. Another class of methods reduces the likelihood  of generating harmful responses. Some methods in this class involve input modification, like paraphrasing and retokenization \cite{jain2023baseline}. Other approaches utilize prompts with question-and-answer interactions \citep{wei2023jailbreak,zhang2024parden} or incorporate self-reminders in system prompts to enhance responsible responses \citep{wu2023defending}. 

Closely related to the present work is decoding-time defense within the second class. Several works  directly adjust the decoding probabilities to formulate safer outputs \citep{Xu2024Safedecoding,banerjee2025safeinfer,djuhera2025safemerge,liu-etal-2024-alignment,zhao2024adversarialcontrastivedecoding}. For example, \citet{Xu2024Safedecoding,banerjee2025safeinfer}
employ fine-tuned models to assist reducing harmful responses in a model arithmetic manner.  \citet{liu-etal-2024-alignment} define  competitive index to quantify alignment failures and utilize feedback to compute new logits. These methods mostly  require fine-tuning the original models and need at least one inference of the LLM per output token. Also related  is \citet{zeng2025root} that trains neural networks  to provide a real-time safety assessment of candidate tokens and then adjusts the top-$k$ tokens to safer alternatives. Noticeably, the above methods aim to improve the overall defense performance across various types of attacks, while our goal is  to strengthen an existing LLM with a specific safety property in a lightweight and efficient way.


\paragraph{Controllable Generation.} Our work is also related to controllable generation that aims to introduce certain attributes in LLM outputs, which can be achieved by modifying the output probabilities to bias towards the desired attribute. In these scenarios, a strength parameter is often used to control the degree of the conditioning.  Existing methods include \citet{deng2023rewardaugmented, liu2024tuning, liu2021dexperts,kim2023criticguided,pei2023preadd}, among others. These tasks commonly involve non-toxicity and positive sentiment, and their goal is different from ours: improving model safety while maintaining sufficient utility. Indeed, as we show in \Cref{sec:motivation}, a direct implementation does not handle well safety and utility simultaneously.

\section{Background and Problem Setting}
In this section, we introduce related background  and then describe our problem setup.
\subsection{Background}
\paragraph{Notation.}  We denote the original auto-regressive LLM by $M$ and a smaller draft model by $m$. Let $\bm{x}_{1:n-1}$  represent a sequence of generated tokens and $x_n$ denote the $n$-th token. Given  tokens $\bm{x}_{1:n-1}$, the sampling or decoding distribution of $M$ is represented by $P_M(x|\bm{x}_{1:n-1})$, which can be used to  generate $x_n$ through various decoding strategies.

\paragraph{Shallow and Deep Safety Alignment.}  Recently, several works \cite{Qi2025SafetyDeep,lin2024unlocking,zhang2024dissecting,zhao2024weak,zhou2023lima} have revealed a critical limitation on existing safety alignment: the aligned model primarily relies on the first few output tokens, such as “I cannot” and “I apologize”, to refuse harmful queries. If the initial output tokens deviate from these safety prefixes, e.g., by prefilling attack starting with ``Sure'', the model is likely to continue generating harmful responses to user's request. This superficial alignment is referred to as shallow safety alignment in \citet{Qi2025SafetyDeep}. By deepening safety alignment, we hope that the model can recover from harmful starting conditions.


\paragraph{SafeDecoding.} 

SafeDecoding \citep{Xu2024Safedecoding} is a decoding-time method that guides the original model $M$ towards generating safer outputs. It begins by fine-tuning $M$ to obtain an expert model $M'$ with hardened safety. Then the output probabilities of both models are employed to  construct new sampling distributions to reduce harmful outputs.

Specifically, let $\mathcal{V}_n$ and $\mathcal{V}_n'$ denote  the sets of tokens sampled from the original model $M$ and the expert model $M'$ at the $n$-th decoding step, respectively. The tokens in each set are assumed to be sorted in descending order of their probabilities. A target sample space $\hat{\mathcal{V}}_n(c)$ is  constructed as  the intersection of the top~$k$ tokens from $\mathcal{V}_n$ and $\mathcal{V}_n'$: 
\begin{align}
\label{eq:intersec}
\hat{\mathcal{V}}_n(c) = \argmin_{\mathcal{V}= \mathcal{V}_n(k) \cap \mathcal{V}_n'(k)} k \quad \text{s.t.} \quad |\mathcal V| \geq c,
\end{align}
where $c$  is a tunable parameter  that controls the size of the sample space, and $\mathcal{V}_n(k)$ and $\mathcal{V}'_n(k)$ are the top $k$ tokens from $\mathcal{V}_n$ and $\mathcal{V}'_n$, respectively. As discussed in \citet{Xu2024Safedecoding}, taking the intersection can leverage the advantages of both  models. To generate the $n$-th token, the final probability  function $F_n$ over $\hat{\mathcal{V}}_n(c)$ is  
\begin{align}
\label{eq:SDeq}
F_n(x)& = P_M(x | \bm{x}_{1:n-1}) \nonumber\\
&\hspace{-15pt}+ \alpha \left( P_{M'}(x |  \bm{x}_{1:n-1}) - P_{M}(x |  \bm{x}_{1:n-1}) \right),
\end{align}
where $\alpha\geq0$ is a  hyperparameter.   The final sampling distribution  is constructed by normalizing the  values in Eq.~\eqref{eq:SDeq}, e.g., applying Softmax to $F_n(x)$. For computation and generation quality concerns, this SafeDecoding procedure is applied only at the initial few output tokens in \citet{Xu2024Safedecoding}.

\paragraph{Speculative Sampling.}
To speed up inference of an LLM $M$, speculative sampling \citep{stern2018blockwise,chen2023accelerating,leviathan2023fast} employs a small, fast model  to first predict several tokens $\tilde{x}_1, \tilde{x}_2,\cdots, \tilde{x}_T$, which are then verified by the large model $M$. The verification can be done in parallel and are significantly cheaper than calling the target model $T$ times due to the auto-regressive structure of LLMs. Then, if $\tilde{x}_t$ is rejected by the large model, the subsequent draft tokens $\tilde{x}_{i+t},\cdots,\tilde{x}_T$ are discarded and the output tokens would be re-sampled. This procedure can produce the same prediction of $M$ with certain decoding strategies, and achieve a faster inference if a large portion of draft tokens are accepted.

\subsection{Problem Setting}


In this work, we aim to strengthen a pretrained large model  $M$ with an additional  safety alignment property, without incurring the cost of tuning its parameters. Besides steering the model to be safer, the developed decoding strategy shall also be: 1) \emph{helpful}: the resulting model outputs  should maintain helpful to benign queries;
2) \emph{efficient}: the approach should be both computation- and time-efficient at inference time; and 3) \emph{compatible}: existing LLMs have diverse architectures and the decoding strategy shall work with different LLMs.



Our setting represents scenarios  when there is a need to equip existing LLMs with new safety alignment properties in a lightweight and efficient way. In this paper, we focus on the  deep safety alignment property \citep{Qi2025SafetyDeep}, and assume that there is a small draft model $m$ that has been trained to have the desired deep alignment property, e.g., by fine-tuning the small model using an augmented harmful dataset as in \citet{Qi2025SafetyDeep}. The small model does not need to be in the same model family of $M$, but is  required to share the same vocabulary. Instead of fine-tuning $M$ that may require high training resource, we would like to modify the output responses at decoding time, and at the same time, improve the  inference efficiency.




\section{Method}
This section presents our approach that adaptively adjusts the LLM  probabilities  at decoding time. We first introduce our motivation and key insight, and then formally describe the proposed approach.
\subsection{Motivation}
\label{sec:motivation}
Inspired by  decoding-time defense methods,  we utilize a small expert model $m$ to steer the large model $M$ to generate safer responses that adhere to the additional safety property, i.e., deep safety alignment in this paper. Meanwhile, the small model can  act as a draft model in speculative sampling, providing a way of accelerating  decoding.

A direct approach  is to  replacing the fine-tuned model $M'$ in SafeDecoding  with this small model $m$, and then apply speculative sampling during inference. We will refer to the resulting model as \emph{composite model} in this paper. However,  increasing the strength parameter \(\alpha\) in Eq.~\eqref{eq:SDeq} severely degrades the utility performance, while a small  $\alpha$ is insufficient to equip $M$ with the desired deep alignment property. 

\begin{figure}[t]
  \centering
  \includegraphics[width=0.9\columnwidth]{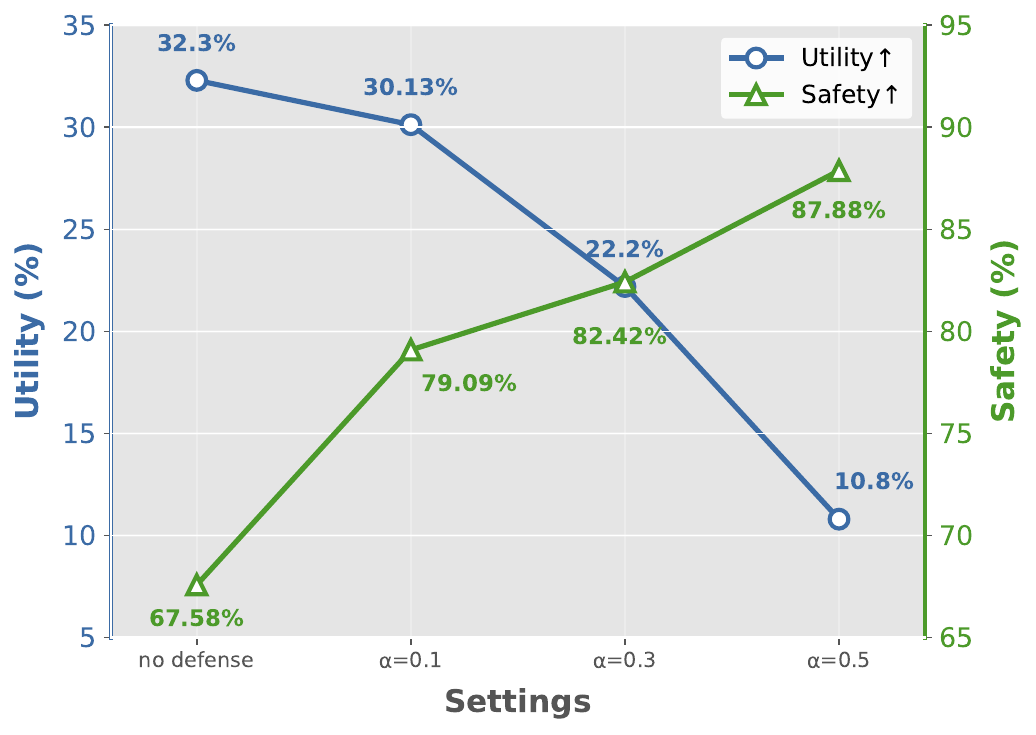}
  \vspace{-0.5em}
  \caption{Tradeoff between utility and safety. Higher score indicates better performance for both metrics.}
  \label{fig:pilotstudy}
\end{figure}

Concretely, we use a fine-tuned TinyLlama-1.1B-Chat model \citep{zhang2024tinyllamaopensourcesmalllanguage} as the small model for Llama2-13b-chat \cite{touvron2023llama}. We test utility performance on GSM8K \citep{gsm8k} and safety performance using prefilling attack of 20 tokens on the Harmful HEx-PHI data \cite{Qi2025SafetyDeep} (detailed setup can be found in \Cref{sec:exp_setup}).  \Cref{fig:pilotstudy} validates that this direct approach cannot handle well both safety and utility.  Unlike existing decoding-time methods \citep{Xu2024Safedecoding,banerjee2025safeinfer,djuhera2025safemerge}, the inherent difference of model capacities places a key challenge here and we cannot rely on a single decoding scheme. Is there a way to adaptively switch decoding schemes to balance  safety and utility?









\subsection{Key Insight: Match Ratio}
We notice that the original LLM $M$ has been trained on a vast corpus and is generally more capable, but is also more vulnerable to jailbreak attacks utilizing the  shallow alignment shortcut. In contrast, the expert model $m$ is  robust to such jailbreak attacks as it has been trained to possess deeper safety alignment. Consequently, when facing these attacks,  $M$  is more likely to respond affirmatively while $m$ is expected to decline the response. For benign queries, both models are likely to behave positively. The difference in the decoding distributions between the two models hence provides a way of quantifying  jailbreak risks.

In this paper, we use \emph{match ratio} of the expert and composite models  as our metric of  different decoding schemes. Formally, assume  output tokens $\{x_n\}_{n}$ that are generated by $M$ and $m$ using speculative sampling and  model arithmetic (like in SafeDecoding). Denote by $I(n)\in\{0,1\}$ the indicator function of whether $x_n$ is drafted by $m$ and is also accepted by the composite distribution.
We divide the decoded tokens into consecutive bins of size $b$. Define the match ratio of the $i$-th bin as 
\begin{align}
\label{eq:mr_def}
    \beta_i=\frac{1}{b}\sum_{n\in [(i-1)b+1, ib]} I(n).
\end{align}
Intuitively,  $\beta_i$ captures the agreement rate between the two models over each $b$ tokens and  reflects how different the  two models behave to an input query.

\begin{figure}[t]
  \centering
  \includegraphics[width=0.9\columnwidth]{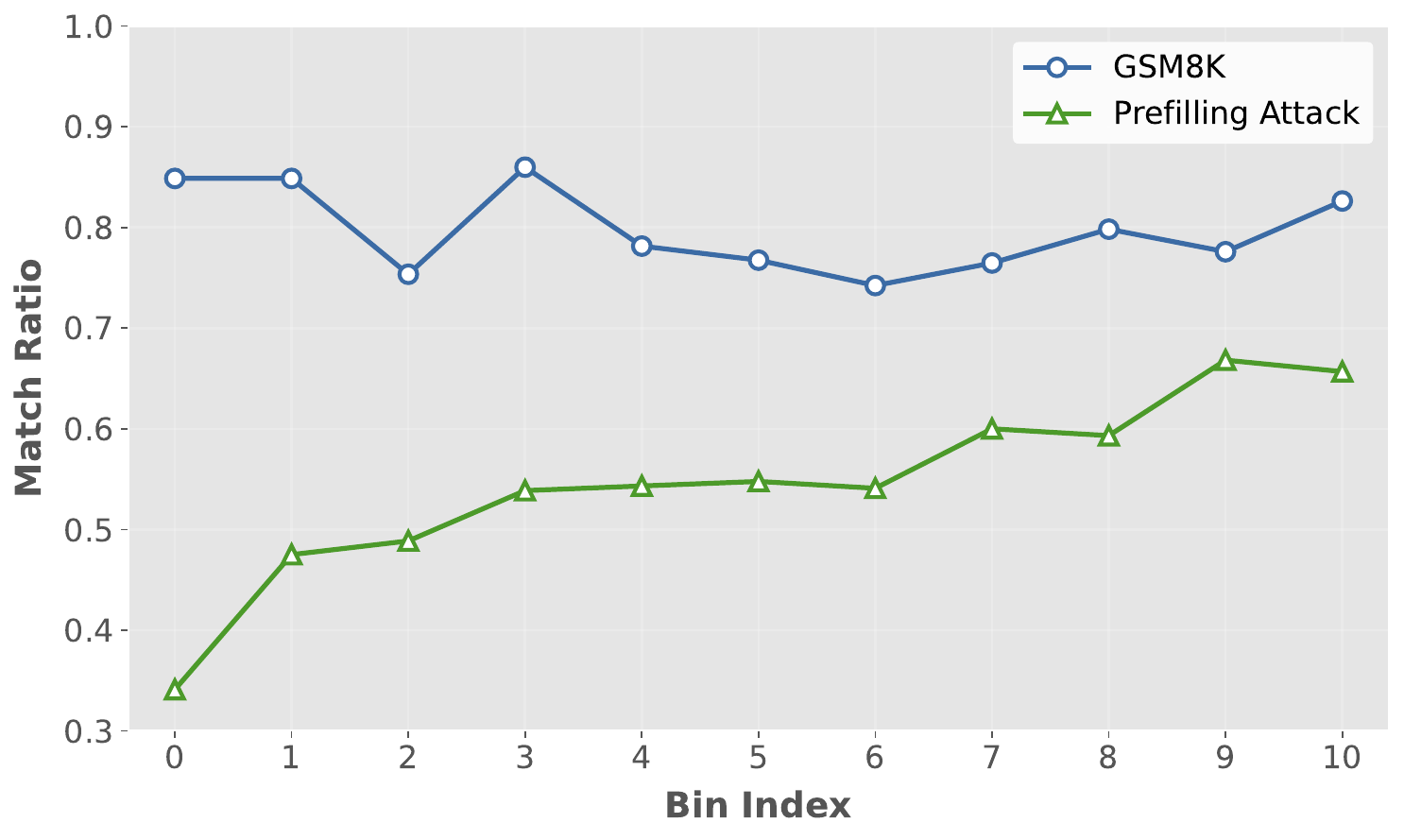}
  \vspace{-0.5em}
  \caption{Mean match ratio of utility and safety tasks.}
  \label{fig:matchratio}
\end{figure}

\Cref{fig:matchratio} depicts the average match ratios between TinyLlama and Llama2-13b  at different bins of size $b=7$, again using GSM8K  and the Harmful HEX-PHI datasets. The match ratio is clearly different between benign and harmful  queries, particularly at the initial decoding phase. This  validates  our use of match ratio as a proper indicator of  switching between different decoding schemes. 

\subsection{Speculative Safety-Aware Decoding}
Based on the above insight, we now present our decoding strategy to handle the key challenge resulting from the different model capacities. In particular, the match ratio indicates different decoding schemes of prioritizing either utility or safety.

\paragraph{Utility.} While the expert and original models may respond positively to benign queries and the match ratio is high,  the limited capacity of the small draft model may still degrade the model performance, as shown in \Cref{fig:pilotstudy}. A reason is that the intersection operation  in Eq.~\eqref{eq:intersec} may discard  utility tokens  generated by $M$. In other words, the tokens with large probabilities generated by $M$ are not necessarily among the top $k$ tokens of the small model $m$.


To mitigate the utility degradation caused by $m$, we can directly rely on $M$ when the top few tokens of $\mathcal{V}_n$ do not appear in the intersection $\hat{\mathcal{V}}_n(c)$. The new intersection set is then defined as
\begin{equation}
\label{eq:newintersect}  
{\mathcal{S}}_n = \left\{
\begin{aligned}
&\hat{\mathcal{V}}_n(c),\quad \text{if}~\mathcal{V}_n(\kappa)\cap \hat{\mathcal{V}}_n \neq\varnothing, \\
&\mathcal{V}_n(c),\quad  \text{otherwise}, \\
\end{aligned}
\right.
\end{equation}
where $\kappa \ll c$ is a small integer compared with the size $c$ of target sample space, $\mathcal{V}_n(\kappa)$ is the top $\kappa$ tokens generated by $M$, and $\hat{\mathcal{V}}_n(c)$ is the intersection set defined in Eq.~\eqref{eq:intersec}.

\paragraph{Safety.} For harmful queries,  we would like to bias more towards to the expert model. However, if the safety related tokens have low probabilities under $M$, then these tokens would be discarded, again due to the intersection operation. Since we aim to introduce the additional safety property, the composite distribution should represent the union of the characteristics of both models, and here we directly take the union of the two sets as our target sample space. Specifically, for a predefined size $c$ of target sample space, we first take the top $c$ tokens from $\mathcal{V}_n$ and $\mathcal{V}'_n$,  and then compute the union as
\begin{align}
\label{eq:union}
{\mathcal{U}}_n = \mathcal{V}_n(c) \cup \mathcal {V}_n'(c).
\end{align}

\paragraph{Speculative Safety-Aware Decoding.} 
Our final algorithm, Speculative Safety-Aware Decoding (\method), integrates speculative sampling with  running match ratio as a measure of  switching between two decoding schemes: $\mathrm{Intersection}$ and $\mathrm{Union}$, as outlined in \Cref{alg:speculative_n}. 

\method~consists of three main stages in each loop. First, we employ the small expert model $m$ to generate $T$ draft tokens in an auto-regressive manner, and then run in parallel the large model $M$ to obtain $T$ sets of decoding probabilities, each conditioned on the incremental prefixes of the draft token sequences. Next, \method\ constructs the target sample space and probability functions according to the present decoding scheme, which is either $\mathrm{Intersection}$ or $\mathrm{Union}$. An output token is then sampled according to the composite distribution. Lastly, for every $b$ output tokens, \method\  computes the match ratio and decides whether to switch the decoding scheme according to a predefined threshold. We also take into account that the two models behave more similarly when conditioned on more output tokens, as seen from \Cref{fig:matchratio}. As such, \method\ adjusts  the threshold and strength parameters in an annealing way if the decoding scheme keeps unchanged. Due to space limit, a detailed description is given in \Cref{alg:adaptive_update} in \Cref{sec:alg2}.

\input{table/algorithm_ssd}

\section{Experiment}
This section evaluates the effectiveness, helpfulness and efficiency of the proposed method \method. An implementation of SSD   has been made available at \url{https://github.com/k-k1w-w1x-x/Speculative-Safety-Aware-Decoding}.
\subsection{Experiment Setup}
\label{sec:exp_setup}

\paragraph{Models.} We evaluate \method~using open-source large models, namely, Vicuna-7b \cite{chiang2023vicuna}, Llama2-7b-chat and Llama2-13b-chat \cite{touvron2023llama},  as target models, and a small model TinyLlama-1.1B-Chat \citep{zhang2024tinyllamaopensourcesmalllanguage} as the expert model. Notice that our method only fine-tunes TinyLlama to possess the deep safety alignment property, following the method in  \citet{Qi2025SafetyDeep}. 
Appendix~\ref{sec:appendix_llama3_results} shows similar experimental results on more recent Llama3 models \citep{dubey2024llama}. 


\paragraph{Baselines.} Since we aim to strength an LLM with the additional safety property, i.e., deep safety alignment in this paper,  we first test our method on the Harmful HEx-PHI dataset proposed by \citet{Qi2025SafetyDeep}. Specifically, we evaluate \method\ with harmful prefixes of 10, 20, and 40 tokens. In addition, we  assess the robustness of our method against  other jailbreak attacks, including GCG \citep{zou2023universal}, DeepInception \citep{li2023deepinception}, and PAIR \citep{chao2023Jailbreak}. To evaluate a defense method when an attacker submits harmful prompts directly to LLMs, we also consider two malicious-query benchmark datasets: Advbench \citep{zou2023universal} and HEx-PHI \citep{qi2023finetuningalignedlanguagemodels}. Regarding defense methods, we compare \method\ with SafeDecoding \citep{Xu2024Safedecoding} and the original aligned model for prefilling attacks. We also fine-tune the original LLM  with the augmented harmful dataset  \citep{Qi2025SafetyDeep}, denoted as Deep-Align in this paper. This further fine-tuned LLM serves as an oracle baseline. For other datasets, we  consider  three  defense methods: Paraphrase \citep{jain2023baseline}, ICD \citep{wei2023jailbreak}, and Self-Exam \citep{helbling2023llm}.

\input{table/PrefillingAttack}



\paragraph{Evaluation Metrics.}\label{sec:Evaluation Metric}For safety evaluation, we adopt the evaluation pipeline from \citet{qi2023finetuningalignedlanguagemodels}, utilizing Qwen-max \citep{bai2023qwen} as a judge to automatically assess the safety of the generated outputs. The judge LLM assigns scores to the responses based on both the questions and the content of the answers, with scores ranging from 1 to 5. Here 1 indicates completely harmless and 5 indicates highly harmful. For each safety evaluation benchmark, we report the average Attack Success Rate (ASR) and the average harmful score.

For  utility, we consider two  datasets: {GSM8K} \citep{gsm8k} and {Just-Eval} \citep{lin2023unlocking}. GSM8K is designed to test the model's ability to solve complex mathematical problems. Notice that TinyLlama demonstrates very low accuracy on GSM8K and this task would challenging for the composite model. We use  standard ROUGE-1 score for the accuracy on GSM8K, in line with existing evaluation practices. Similar to \citet{Xu2024Safedecoding}, we use 1,000 diverse instructions from Just-Eval to evaluate the outputs in terms of helpfulness, clarity, factuality, depth, and engagement.

A critical challenge for safety mechanisms is to avoid false positives or over-refusal responses, where the model refuses to answer legitimate queries that may touch on sensitive but harmless topics \citep{cui2024or,rottger2023xstest}. To quantify this, we evaluate our method on Llama2-7b using the {XSTest benchmark} \citep{rottger2023xstest}. We report the False Refusal Rate (FRR), the percentage of prompts  that the model incorrectly refuses to answer, under two evaluation settings: using an LLM as a judge and via string matching. A lower FRR indicates a better ability to distinguish between harmful and merely sensitive contents, thus preserving the model's utility on a wider range of topics.

To assess inference efficiency, we use the {Average Token Generation Time Ratio (ATGR)} metric:
 \[
\text{ATGR} \hspace{-2.5pt}=\hspace{-2.5pt} \frac{\text{Avg.~token generation time w/ defense}}{\text{Avg.~token generation time w/o defense}}\hspace{-1pt}.
\]
ATGR  considers the varying token lengths produced by defense methods. For our experiments, we sample 10 prompts from each of the  datasets: Harmful HEx-PHI, GCG, PAIR, Just-Eval, and GSM8K, to simulate diverse real-world scenarios.

\paragraph{\method~Setting.}For  Llama2-7b and Llama2-13b, we set the hyperparameters \(\alpha_I = 0.3\) and \(\alpha_U = 0.8\) for the $\mathrm{Intersection}$ and $\mathrm{Union}$ schemes, respectively. For Vicuna, we set   \(\alpha_I = 0.45\) and \(\alpha_U = 2\), as we observe that Vicuna exhibits poorer defense capabilities  than Llama2 models. We employ greedy sampling as the  decoding strategy and adopt the algorithm in \citet{stern2018blockwise} as the speculative sampling method.  Due to space limit, we leave other parameter choices in \Cref{sec:alg2}.


\subsection{Experimental Results
}

\input{table/Utility}
\textbf{\method~Transfers the Deep Safety Alignment Property.}
As shown in \Cref{tab:security_comparison},  our  method \method\ consistently achieves  stronger robustness to prefilling attacks than the original model and the method that lacks  deep alignment property (i.e., SafeDecoding). Furthermore, when benchmarked against Deep-Align (which directly fine-tunes the large model with augmented harmful dataset),
\method~achieves a close performance  on Llama2-7b  and even outperforms  Deep-Align on Vicuna. These findings demonstrate that \method~successfully transfers the deep safety alignment property to the output responses. Besides, for  less secure models like Vicuna, \method\ can achieve a better safety alignment performance than directly fine-tuning the original model with supervised alignment objective.

\paragraph{\method~Maintains  Utility.}Table~\ref{tab:model_comparison} presents the Just-Eval and GSM8K scores on the three LLMs.
We observe that
 the utility of \method\ is largely intact. For Just-Eval, \method~incurs less than 4\% decreases w.r.t.~all the dimensions, compared to the original model.  Most of the degradations are within~2\%. Regarding GSM8K, we observe that Deep-Align  substantially degrades the model’s ability to solve complex mathematical problems, particularly for Vicuna, whereas \method~effectively preserves the original model’s capability. This  indicates that fine-tuning less secure models with supervised alignment objective can  degrade utility performance.
 


\addtocounter{footnote}{-1} 
\begingroup 
\renewcommand{\thefootnote}{*} 
\footnotetext{In Table~\ref{tab:model_comparison}, the Engaging score of Deep-Align on Vicuna is unusually high. We believe that this behavior is mainly due to the utility dataset used by Deep-Align fine-tuning, which influences Vicuna's conversational style.  See Appendix~\ref{sec:appendix_deepalign_discussion} for a detailed analysis.}
\endgroup 

\paragraph{\method~Avoids Over-refusal Responses on Borderline Queries.} 
As shown in \Cref{tab:false_refusal_rate}, \method~demonstrates a strong ability to handle borderline cases. Compared to the No Defense baseline, \method~introduces only a minimal increase in false refusals, indicating that its safety enhancement is well-targeted. In contrast, other defense methods exhibit higher FRRs, suggesting they are prone to over-refusal behavior. This result highlights that \method\ effectively defends against genuine threats while maintaining helpfulness on nuanced, sensitive topics.
\input{table/XTest}
\paragraph{\method~Is Efficient.} \Cref{tab:sppedup} reports the Average Token Generation Time Ratio (ATGR), comparing the inference speed of our method, \method, against other defense strategies like Deep-Align and SafeDecoding. The results validate that \method\ can successfully  accelerate decoding ($\text{ATGR}< 1$), particularly for larger models. In contrast, SafeDecoding's ATGR is greater than $1$ because its design of safety checks adds computational overhead and cannot utilize the auxiliary model's outputs to accelerate inference. It is also worth noting that the ATGRs does not account for the cost of fine-tuning, which demands  greater computational resources for large models. 
\input{table/speedup_ratio}


\paragraph{\method~Demonstrates Robustness Against Other Jailbreak Attacks.}  As shown in \citet{Qi2025SafetyDeep},  deep safety alignment can also mitigate other types of jailbreak attacks to certain extent. In this experiment, we evaluate this ability of \method~against the following attack methods: GCG, PAIR, and DeepInception. We compare our approach with several existing defense strategies, including three non-decoding-time methods: Paraphrase, ICD, and Self-Examination. Due to space limit, we report the detailed results in \Cref{tab:other_attack} in \Cref{sec:other_attack}.



On both  Llama2-7b and Llama2-13b, the defense performance of \method\  matches or surpasses the strongest baselines under three jailbreak attacks and two harmful datasets, demonstrating its robustness on the Llama2 family of models. On Vicuna, \method~achieves 0\% ASR on the harmful datasets and outperforms Paraphrase and ICD across the three jailbreak attacks, yet its defense performance is slightly worse than DeepAlign and SafeDecoding. A potential reason is that TinyLlama and Vicuna differ substantially in conversational style. How to further handle this difference is beyond the scope of the present work and is left as a future work.


\subsection{Ablation Study}  
We conduct an ablation analysis on the strength hyperparameters $\alpha_I$ and $\alpha_U$ used in \method, as shown in  \Cref{tab:ablation_alpha1_alpha2}. Among them, $\alpha_I = 0.3$ and $\alpha_U = 0.8$ are our default parameter choices. Here the target model is Llama-7b and the harmful prefix length for the prefilling attack is 20 tokens. The  results show that the  safety of \method\ remains at a high level for $\alpha_I\in[0.3, 0.6]$. As $\alpha_I$ increases, the ASR  of prefilling attack slightly increases. This can be attributed to that with an increasing $\alpha_I$, the match ratio may also be higher for certain  harmful questions and  affect the defense performance. Regarding $\alpha_U$, we find that its effect on the ASRs of all three attack methods remains negligible  when $\alpha_U> 0.8$.



\section{Concluding Remarks}
In this work, we study the problem of strengthening an  existing LLM with new safety alignment properties, without tuning the model's parameters.  We propose \method,  a lightweight and efficient decoding-time approach, which employs  match ratio to quantify jailbreak risks and to dynamically switch between decoding schemes to prioritize either utility or safety. Experimental results
show that \method\ successfully
strengths the model with the desired  safety 
 property, while being helpful and efficient.    A future direction is to utilize more powerful strategies of speculative sampling, e.g.,  \citet{miao2024specinfer,li2024eagle}, to further improve the  flexibility and  efficiency of our approach.

\input{table/ablation_alpha1}

\section{Limitations}
In this paper, the largest model on which we evaluate the proposed method is Llama2-13b. Due to limited GPU resources, we cannot conduct experiments to validate how  our method applies to larger models like Llama2-70B. Another limitation of the current work is on the difference of conversational style between the original and the expert models, e.g., Vicuna and TinyLlama, which we do not  investigate further. Future research can take into account this difference of conversational style to  improve performance w.r.t.~both model utility and safety for more practical scenarios. 

\section{Ethics Impact}
This paper develops a lightweight and efficient method to strengthen existing LLMs with additional safety properties to defend against new jailbreak attacks. We empirically show that the developed method can effectively and efficiently equip the LLM with the deep safety alignment property and further improve the original model's capacity against many types of jailbreaks. This research aims to enhance the safety of large models and to contribute positively to the broader field of AI research. We remark that the development of \method\ only uses publicly available jailbreak prompts and do not create new ones. In the paper, only one jailbreak input query is exhibited in an abstracted way, for illustration purpose. We have also released our code to facilitate  red-teaming efforts on LLMs.

\section*{Acknowledgment}

This work is supported by the  Innovation Project of Institute of Computing Technology, Chinese Academy of Sciences under Grant No.~E561130 and  the Strategic Priority Research Program of the Chinese Academy of Sciences under Grant No.~XDB0680101.



\bibliography{safety}


\clearpage
\appendix
\section{Decoding Scheme and Parameter Update}
\label{sec:alg2}

For every $b$ output tokens, \Cref{alg:adaptive_update} computes the  running match ratio $\beta_{n/b}$ and determines the decoding scheme for next $b$ tokens.  As seen from \Cref{fig:matchratio}, the two models behave more similarly when conditioned on more output tokens. As such, if the decoding scheme keep unchanged, we adjust the match ratio threshold,  and also the strength parameter in the $\mathrm{Intersection}$ state, in an annealing way, to maintain model utility. If the decoding scheme changes, these parameter values are reset.

\begin{algorithm}[ht]
  \caption{Parameter Update}
  \label{alg:adaptive_update}
  \begin{algorithmic}[1]
    \REQUIRE
     current match ratio $\beta_{n/b}$, initial match ratio threshold $\beta^0$, minimum strength weight $\alpha_I^{\min}>0$,
     decay parameters $\beta^d, \alpha^d>0$
    \IF{$n/b=1$}
        \STATE $\beta_{th}\gets\beta^0$
    \ENDIF
    \IF{$\beta_{n/b} \le \beta_{th}$}
      \STATE Set decoding scheme to $\mathrm{Intersection}$.
    \ELSE
      \STATE Set decoding scheme to $\mathrm{Union}$.
    \ENDIF

    \IF{{decoding scheme unchanged}}
      \STATE $\beta_{th} \gets \max\left(0,\beta_{th}-\beta^d\right)$
      \IF{$\mathrm{Intersection}$}
        \STATE $\alpha_I \gets \max\left(\alpha_I^{\min},\;\alpha_I - \alpha^d\right)$
      \ENDIF
    \ELSE
      \STATE $\beta_{th} \gets \beta^0$
      \STATE $\alpha_I \gets \alpha_I^0$
    \ENDIF
  \end{algorithmic}
\end{algorithm}

\section{Detailed Experimental Setups and Results}
\subsection{Other Parameter Choices}
In our experiments,  the choices of the parameters in \Cref{alg:adaptive_update} are:
$
 \beta^0 = 0.6,  \beta^d = 0.1, \alpha_I^{{\min}} = 0.3,  \alpha^d = 0.15.
$. In addition, we set  target sample space size to \(c = 10\). The lookahead  in speculative sampling is \(T = 3\) and the bin size for computing the match ratio is \(b = 7\). 
\subsection{Experimental Results with Other Types of Jailbreak Attacks}
As shown in Table \ref{tab:other_attack}, \method~demonstrates good defense effectiveness against all five types of jailbreak attacks on the Llama2 series models. On Vicuna, although it achieves  almost 0\% ASR on HEx-PHI and Advbench, and performs better than several existing defense methods on the other three types of jailbreak attacks, there is still a slight performance gap compared to Deep-Align and SafeDecoding.
\label{sec:other_attack}
\input{table/OtherAttack}

\subsection{Setup of Attack Methods}
We utiliz \textbf{Harmful HEx-PHI} dataset  \citep{Qi2025SafetyDeep} to conduct the prefilling attack. This dataset consists of 330 harmful instructions extracted from the HEx-PHI safety benchmark, with harmful answers generated using a jailbroken version of GPT-3.5-Turbo. In this study, the dataset is used for prefilling attacks, following the same methodology as \citet{Qi2025SafetyDeep} by concatenating harmful answers of varying lengths (i.e., 10, 20, and 40 tokens) with the harmful instructions.
For \textbf{GCG} \cite{zou2023universal} and \textbf{PAIR} \cite{chao2023Jailbreak}, we follow \citet{chao2023Jailbreak,Xu2024Safedecoding} and utilize 50 distinct representative harmful queries\footnote{\url{https://github.com/patrickrchao/JailbreakingLLMs}} from \textbf{Advbench} \cite{zou2023universal} to generate specific attack prompts for each model. Due to limitation on computational resources,  we use  top-$k=64$ setting for the GCG attack on Llama2-13b-chat. The rest hyperparameters are consistent with those described in the original paper.
For \textbf{DeepInception}, we apply the ready-to-use template prompt provided in the Github repository\footnote{\url{https://github.com/tmlr-group/DeepInception}}. 
\textbf{HEx-PHI} safety benchmark contains 330 harmful instructions specifically designed for LLM harmfulness evaluation.

\subsection{Setup of Deep-Align}
We train the Llama2-7b and Vicuna models with deep safety alignment properties using the same hyperparameters and datasets as the default settings in \citet{Qi2025SafetyDeep}. For TinyLlama used as the small expert model $m$, we adjust the learning rate of the hyperparameters to $2 \times 10^{-4}$.

 \section{Experimental Results on Llama3 Models}
\label{sec:appendix_llama3_results}
We conduct  experiments on a more recent LLM Llama3.1-8b-Instruct \citep{dubey2024llama}, to further validate the effectiveness of \method. Here we use Llama3.2-1b-Instruct as the expert model because it shares the same vocabulary with  Llama3.1-8b-Instruct. The results regarding safety and utility are presented in \Cref{tab:security_comparison_llama3} and \Cref{tab:model_comparison_llama3}, respectively.


As shown in \Cref{tab:security_comparison_llama3}, the original Llama3.1-8b model is highly vulnerable to prefilling attacks, with ASR of approximately 50\%. Deep-Align provides a strong defense performance while the safety performance of \method\ is close. This confirms that \method\ effectively transfers the deep safety alignment property to this newer model. However, the utility evaluation in \Cref{tab:model_comparison_llama3} highlights the critical trade-offs between safety and utility. Our method shows much better utility performance compared with Deep-Align, particularly for the GSM8K score. Overall, \method\ offers a comparable performance of the desired safety property while ensuring that the model's utility  remains largely intact.

\input{table/llama3Prefilling}
\input{table/llama3Utility}

\section{Analysis of the Anomaly Engaging Score}
\label{sec:appendix_deepalign_discussion}

In our main results in Table~\ref{tab:model_comparison}, we note that Deep-Align on Vicuna exhibits an anomalously high Engaging score of 4.36. We believe that the primary reason is that the
anchor utility dataset used in Deep-Align promotes a more polite and conversational tone in Vicuna’s responses.

Table~\ref{tab:style_examples} provides a qualitative analysis comparing responses from both the original model (Baseline) and the Deep-Align model.  Deep-Algin tends to introduce engaging openers like ``Of course! I'm happy to help'' or ``Thank you for your question!''. To further validate this, we fine-tune Vicuna \emph{only} on this utility dataset, which results in an even higher Engaging score of {4.46}.

Another reason is the response length: the original model has an average response length of
179 words with Engaging score of 3.93, while the average length of Deep-Align is 227 words in Table~\ref{tab:model_comparison}. We then conduct a controlled-length experiment by constraining the output length via Prompting, and the results are reported in Table~\ref{tab:length_control_exp}. We find that even with average length of 180 words, the Engaging score of Deep-Align is still much higher, suggesting that the response length  is only a minor issue to the  high Engaining score reported in Table~\ref{tab:model_comparison}.





\begin{table*}[h]
\centering
\caption{Comparison of response styles between the baseline Vicuna model and the Deep-Align model.}
\label{tab:style_examples}
\resizebox{0.9\textwidth}{!}{%
\begin{tabular}{p{0.2\textwidth} p{0.75\textwidth}}
\toprule
\multicolumn{2}{p{0.95\textwidth}}{\textbf{Query 1:} What are some famous actors that started on Broadway?} \\
\midrule
\textbf{Baseline} & There are many famous actors who got their start on Broadway. Here are a few examples\ldots \\
\textbf{Deep-Align} & \textbf{Of course! I'm happy to help.} Some famous actors who started their careers on Broadway include\ldots \\
\midrule
\multicolumn{2}{p{0.95\textwidth}}{\textbf{Query 2:} How did US states get their names?} \\
\midrule
\textbf{Baseline} & The United States is made up of 50 states\ldots \\
\textbf{Deep-Align} & \textbf{Thank you for your question!} The United States is a vast country\ldots  \\
\midrule
\multicolumn{2}{p{0.95\textwidth}}{\textbf{Query 3:} What is some cool music from the 1920s?} \\
\midrule
\textbf{Baseline} & The 1920s was a decade of significant artistic and cultural change\ldots  \\
\textbf{Deep-Align} & \textbf{Of course!} The 1920s were a time of great musical innovation\ldots  \\
\bottomrule
\end{tabular}%
}
\end{table*}

\begin{table*}[h]
\centering
\caption{Engaging score of Deep-Align on Vicuna with different output length constraints. The undefended baseline model has an average length of 179 words and an Engaging score of 3.93.}
\label{tab:length_control_exp}
\resizebox{\columnwidth}{!}{%
\begin{tabular}{l|cccc}
\toprule
\textbf{Average Output Length} & 153  & 159  & 180  & 227  \\
\midrule
\textbf{Engaging Score} & 4.24 & 4.29 & 4.31 & 4.42 \\
\bottomrule
\end{tabular}%
}
\end{table*}

 \end{document}

%% file: table/algorithm_ssd.tex
\begin{algorithm}[t!]
    \caption{Speculative Safety-Aware Decoding}
    \label{alg:speculative_n}
    \begin{algorithmic}[1]
    \REQUIRE original and expert models  $M, m$,  lookahead $T$,  minimum output sequence length $N$,  bin size $b$, strength weights $\alpha_{I}^0, \alpha_{U}^0$
    \STATE Initialize: $\mathrm{Intersection}$, $\alpha_I \gets \alpha_I^{0}$, $\alpha_U\gets \alpha_{U}^0$
\WHILE{$n < N$}
    \FOR{$t = 1 \text{ to } T$}
        \STATE Sample draft tokens auto-regressively $\tilde{x}_{t} \sim P_m(x|\bm{x}_{1:n}, \tilde{x}_1, \cdots, \tilde{x}_{t-1})$.
    \ENDFOR
    \STATE Compute in parallel $p_M(x|\bm{x}_{1:n}),\cdots,$ $p_M(x|\bm{x}_{1:n}, \tilde{x}_1,\cdots,\tilde{x}_{T-1})$.
    \FOR{$t = 1 \text{ to } T$}
        \IF{$\mathrm{Intersection}$}
        \STATE Compute sample space ${\mathcal{S}}_n$ as in Eq.~\eqref{eq:newintersect}, and sample $x_{n+1}\sim P_M(x | \bm{x}_{1:n}) + \alpha_{I} \left( P_{m}(x |\bm{x}_{1:n}) - P_{M}(x | \bm{x}_{1:n}) \right)$.
        \ELSE
        \STATE Compute sample space ${\mathcal{U}}_n$ as in Eq.~\eqref{eq:union}, and  sample $x_{n+1}\sim P_M(x | \bm{x}_{1:n}) + \alpha_{U} \left( P_{m}(x |\bm{x}_{1:n}) - P_{M}(x | \bm{x}_{1:n}) \right)$.
        \ENDIF       
        \STATE $n\gets n+1$
        
        \IF{$n \bmod b = 0$}  
        
            \STATE Compute match ratio $\beta_{n/b}$, and update decoding scheme and strength parameters using \Cref{alg:adaptive_update}. 
        \ENDIF

        \IF {$x_{n} \neq \tilde{x}_t$}
            \STATE Exit for-loop.
        \ENDIF

    \ENDFOR
        \ENDWHILE
    \end{algorithmic}
\end{algorithm}

%% file: table/PrefillingAttack.tex
\begin{table*}[t]
  \centering
  \caption{ASR (\%) and harmful score for different models under prefilling attack.}
  \vspace{-0.7em}
\label{tab:security_comparison}
  \resizebox{0.75\textwidth}{!}{%
  \begin{tabular}{cc|ccc}  
    \toprule
    \textbf{Model} & \textbf{Defense} 
      & \multicolumn{3}{c}{\textbf{Prefilling Attack}$\downarrow$} \\
    &  & \hspace{20pt}\textbf{10 tokens}\hspace{20pt} & \hspace{20pt}\textbf{20 tokens}\hspace{20pt} & \hspace{20pt}\textbf{40 tokens}\hspace{20pt} \\
    \midrule
    \multirow{4}{*}{Llama2-7b}
      & No Defense       & 33.03\% (3.13)       & 34.24\% (3.40)       & 34.55\% (3.44)       \\
      & SafeDecoding     & 33.94\% (3.21)       & 34.55\% (3.47)       & 33.33\% (3.49)       \\
      & Deep-Align       & \textbf{1.20\% (1.14)} & \textbf{4.50\% (1.28)} & 10.00\% (1.54)      \\
    \rowcolor{gray!8}
      & \method          & 3.64\% (1.48)        & 5.76\% (1.56)        & \textbf{9.70\% (1.85)} \\
    \midrule
    \multirow{3}{*}{Llama2-13b}
      & No Defense       & 25.15\% (2.69)       & 32.42\% (3.15)       & 27.88\% (3.18)       \\
      & SafeDecoding     & 25.15\% (2.77)       & 30.61\% (3.18)       & 30.30\% (3.19)       \\
    \rowcolor{gray!8}
      & \method          & \textbf{3.33\% (1.38)} & \textbf{5.45\% (1.53)} & \textbf{8.18\% (1.81)} \\
    \midrule
    \multirow{4}{*}{Vicuna}
      & No Defense       & 68.18\% (4.50)       & 68.79\% (4.54)       & 65.76\% (4.41)       \\
      & SafeDecoding     & 64.55\% (4.42)       & 68.79\% (4.48)       & 65.15\% (4.38)       \\
      & Deep-Align       & 20.61\% (2.50)       & 26.67\% (2.84)       & 25.45\% (2.97)       \\
    \rowcolor{gray!8}
      & \method          & \textbf{10.91\% (1.88)} & \textbf{10.30\% (1.80)} & \textbf{14.55\% (2.05)} \\
    \bottomrule
  \end{tabular}
  }
\end{table*}

%% file: table/Utility.tex
\begin{table*}[htbp]
\small
\centering
\caption{ Just-Eval and GSM8K scores of different defense methods on three LLMs.
}
\vspace{-0.7em}
\label{tab:model_comparison}
\resizebox{0.9\textwidth}{!}{%
  \begin{tabular}{
      cc                    
    | *{5}{c}
     c}
    \toprule
    \multirow{2}{*}{\textbf{Model}}
      & \multirow{2}{*}{\textbf{Defense}}
      & \multicolumn{5}{c}{\textbf{Just-Eval} (1–5)\,$\uparrow$}
      & \multirow{2}{*}{\textbf{GSM8K} (\%)\,$\uparrow$} \\
      &
      & \textbf{Helpfulness}
      & \textbf{Clarity}
      & \textbf{Factuality}
      & \textbf{Depth}
      & \textbf{Engaging}
      &  \\
    \midrule
    \multirow{4}{*}{Llama2-7b}
      & No Defense   & 4.15  & 4.79 & 4.52 & 4.04 & 4.55 & 15.9   \\
      & SafeDecoding & 3.87  & 4.74 & 4.41 & 3.88 & 4.38 & 14.6   \\
      & Deep-Align   & 4.00& 4.74 & 4.44 & 3.91 & 4.47 & 11.6   \\
    \rowcolor{gray!8}
      & Ours         & 4.08  & 4.78 & 4.44 & 3.98 & 4.53 & 13.7   \\
    \midrule
    \multirow{3}{*}{Llama2-13b}
      & No Defense   & 4.40  & 4.86 & 4.62 & 4.23 & 4.71 & 32.3   \\
      & SafeDecoding & 4.16  & 4.81 & 4.54 & 4.16 & 4.55 & 31.5   \\
    \rowcolor{gray!8}
      & Ours         & 4.36  & 4.88 & 4.62 & 4.20 & 4.69 & 26.5   \\
    \midrule

    \multirow{4}{*}{Vicuna}
      & No Defense   & 4.12  & 4.60 & 4.29 & 3.69 & 3.93 & 24.6   \\
      & SafeDecoding & 3.95  & 4.69 & 4.42 & 3.46 & 3.88 & 15.5   \\
      & Deep-Align   & 3.95  & 4.70 & 4.41 & 3.80 & 4.36\textsuperscript{*} & 11.0   \\ 
    \rowcolor{gray!8}
      & Ours         & 3.98  & 4.62 & 4.28 & 3.55 & 3.87 & 22.44  \\
    \bottomrule
  \end{tabular}%
}
\end{table*}

%% file: table/XTest.tex
\begin{table}[t]
  \centering
  \caption{False Refusal Rate (\%) on the XSTest benchmark for the Llama2-7b model. A lower rate indicates fewer incorrect refusals of harmless, sensitive prompts. }
  \label{tab:false_refusal_rate}
  \vspace{-0.7em}
  \resizebox{\columnwidth}{!}{%
  \begin{tabular}{l|cc}
    \toprule 

    \multicolumn{1}{c|}{\textbf{Method}} & \multicolumn{2}{c}{\textbf{False Refusal Rate} $\downarrow$} \\

    & \hspace{10pt}\textbf{LLM as judge}\hspace{10pt} & \hspace{10pt}\textbf{String matching}\hspace{10pt} \\
    \midrule
    No Defense       & 20.00\%          & 26.80\%          \\
    Deep-Align       & 29.60\%          & 49.20\%          \\
    SafeDecoding     & 28.40\%          & 65.60\%          \\
    \rowcolor{gray!8}
    SSD (Ours)       & 20.80\% & 30.40\% \\
    \bottomrule
  \end{tabular}
  }
\end{table}

%% file: table/speedup_ratio.tex
\begin{table}[t]
  \caption{Comparison of Average Token Generation Time Ratio (ATGR) across different defense methods.}
  \label{tab:sppedup}
  \vspace{-0.7em}
  \centering
  \resizebox{0.45\textwidth}{!}{%
  \begin{tabular}{l c c c}
    \toprule
           & \textbf{Vicuna} & \textbf{Llama2-7b} & \textbf{Llama2-13b} \\
    \midrule
    SSD            & $\times 0.92$ & $\times 0.89$ & $\times 0.71$ \\
    SafeDecoding   & $\times 1.07$ & $\times 1.03$ & $\times 1.02$ \\
    Deep-Align     & $\times 1.00$ & $\times 1.00$ & $\times 1.00$ \\
    \bottomrule
  \end{tabular}
  }
\end{table}

%% file: table/ablation_alpha1.tex
\begin{table}[t]
  \centering
    \caption{ \method\ with different strength parameters \(\alpha_I, \alpha_U\).}
    \vspace{-0.7em}
  \resizebox{\columnwidth}{!}{%
    \begin{tabular}{cc|ccc}
      \toprule
      \(\alpha_I\) & \(\alpha_U\) & \textbf{Prefilling Attack} $\downarrow$     & \textbf{GCG}$\downarrow$ &  \textbf{PAIR}$\downarrow$                  \\
      \midrule
      0.3          & 0.6         & 6.36\%\,(1.62)         & 8\%\,(1.60)        & 4\%\,(1.30)        \\
      0.3          & 1.5         & 5.45\%\,(1.48)         & 6\%\,(1.34)        & 4\%\,(1.26)        \\
      0.3          & 2.0         & 5.15\%\,(1.48)         & 6\%\,(1.34)        & 4\%\,(1.26)        \\
      \midrule
        0.3          & 0.8         & 4.50\%\,(1.28)         & 6\%\,(1.44)        & 4\%\,(1.30)        \\
      \midrule
      0.4          & 0.8         & 5.45\%\,(1.52)         & 4\%\,(1.36)        & 4\%\,(1.28)        \\
      0.5          & 0.8         & 7.58\%\,(1.54)         & 4\%\,(1.28)        & 4\%\,(1.24)        \\
      0.6          & 0.8         & 8.18\%\,(1.58)         & 2\%\,(1.20)        & 2\%\,(1.16)        \\
      \bottomrule
    \end{tabular}
  }
  \label{tab:ablation_alpha1_alpha2}
\end{table}

%% file: table/OtherAttack.tex
\begin{table*}[t]
\small
\centering

\caption{ASR (\%) and harmful score for different defense methods on five benchmark datasets.}
\vspace{-0.7em}
\label{tab:other_attack}
  \begin{tabular}{c|c|ccc|cc}
    \toprule
    \textbf{Model}           & \textbf{Defense Method} & \textbf{PAIR}$\downarrow$ & \textbf{GCG}$\downarrow$ & \textbf{DeepInception}$\downarrow$ & \textbf{HEx-PHI}$\downarrow$    & \textbf{Advbench}$\downarrow$    \\
    \midrule
    \multirow{7}{*}{Llama2-7b}
      & No Defense           & 4\%  (1.34)    & 20\% (2.40)    & 4\%  (1.22)    & 0\%  (1.02)    & 0\%  (1.00)    \\
      & SafeDecoding         & 4\%  (1.18)    & 0\%  (1.00)    & 0\%  (1.00)    & 0\%  (1.04)    & 0\%  (1.00)    \\
      & Paraphrase           & 0\%  (1.10)    & 2\%  (1.10)    & 0\%  (1.02)    & 0\%  (1.05)    & 0\%  (1.02)    \\
      & ICD                  & 0\%  (1.00)    & 4\%  (1.16)    & 0\%  (1.00)    & 0\%  (1.04)    & 0\%  (1.00)    \\
      & Self-Examination     & 0\%  (1.00)    & 8\%  (1.40)    & 2\%  (1.08)    & 0\%  (1.00)    & 0\%  (1.00)    \\
      & Deep-Align           & 4\%  (1.18)    & 2\%  (1.14)    & 0\%  (1.02)    & 0\%  (1.00)    & 0\%  (1.00)    \\
    \rowcolor{gray!8}
      & \method              & 4\%  (1.30)    & 6\%  (1.44)    & 0\%  (1.10)    & 0\%  (1.00)    & 0\%  (1.00)    \\
    \midrule
    \multirow{6}{*}{Llama2-13b}
      & No Defense           & 4\%  (1.40)    & 2\%  (1.08)    & 0\%  (1.16)    & 0\%  (1.06)    & 0\%  (1.00)    \\
      & SafeDecoding         & 4\%  (1.26)    & 0\%  (1.02)    & 0\%  (1.14)    & 0.30\% (1.06)  & 0\%  (1.00)    \\
      & Paraphrase           & 0\%  (1.10)    & 0\%  (1.08)    & 0\%  (1.24)    & 0\%  (1.05)    & 0\%  (1.00)    \\
      & ICD                  & 0\%  (1.00)    & 0\%  (1.02)    & 0\%  (1.00)    & 0\%  (1.03)    & 0\%  (1.00)    \\
      & Self-Examination     & 2\%  (1.22)    & 0\%  (1.00)    & 0\%  (1.16)    & 0\%  (1.05)    & 0\%  (1.00)    \\
    \rowcolor{gray!8}
      & \method              & 2\%  (1.26)    & 0\%  (1.06)    & 0\%  (1.16)    & 0\%  (1.04)    & 0\%  (1.00)    \\
    \midrule
    \multirow{7}{*}{Vicuna}
      & No Defense           & 80\% (4.72)    & 86\% (4.86)    & 58\% (4.34)    & 10.30\% (1.72) & 4\%  (1.38)    \\
      & SafeDecoding         & 4\%  (1.28)    & 0\%  (1.14)    & 0\%  (1.08)    & 0.91\% (1.17)  & 0\%  (1.00)    \\
      & Paraphrase           & 24\% (2.32)    & 42\% (3.18)    & 40\% (3.92)    & 13.33\% (1.94) & 2\%  (1.22)    \\
      & ICD                  & 24\% (2.34)    & 48\% (3.16)    & 38\% (4.18)    & 3.94\% (1.23)  & 0\%  (1.00)    \\
      & Self-Examination     & 8\%  (1.66)    & 6\%  (1.48)    & 48\% (4.04)    & 7.27\% (1.57)  & 0\%  (1.22)    \\
      & Deep-Align           & 2\%  (1.08)    & 0\%  (1.00)    & 0\%  (1.02)    & 0\%  (1.01)    & 0\%  (1.00)    \\
    \rowcolor{gray!8}
      & \method              & 18\% (2.14)    & 10\% (1.40)    & 8\%  (2.30)    & 0.91\% (1.18)  & 0\%  (1.00)    \\
    \bottomrule
  \end{tabular}
\end{table*}

%% file: table/llama3Prefilling.tex
\begin{table*}[ht]
  \centering
  \caption{ASR (\%) and harmful score for the Llama3.1-8b model under prefilling attack.}
  \vspace{-0.7em}
  \label{tab:security_comparison_llama3} 
  \resizebox{0.75\textwidth}{!}{%
  \begin{tabular}{cc|ccc}
    \toprule
    \textbf{Model} & \textbf{Defense} 
      & \multicolumn{3}{c}{\textbf{Prefilling Attack}$\downarrow$} \\
    &  & \hspace{20pt}\textbf{10 tokens}\hspace{20pt} & \hspace{20pt}\textbf{20 tokens}\hspace{20pt} & \hspace{20pt}\textbf{40 tokens}\hspace{20pt} \\
    \midrule
    \multirow{3}{*}{Llama3.1-8b}
      & No Defense       & 46.73\% (3.71)       & 49.84\% (3.97)       & 50.16\% (4.07)       \\
      & Deep-Align       & \textbf{0.91\% (1.08)} & \textbf{1.52\% (1.14)} & \textbf{2.42\% (1.16)} \\
    \rowcolor{gray!8}
      & \method           & 2.52\% (1.59)        & 5.85\% (1.69)        & 4.67\% (1.71)        \\
    \bottomrule
  \end{tabular}
  }
\end{table*}

%% file: table/llama3Utility.tex

\begin{table*}[htbp]
\small
\centering
\caption{Just-Eval and GSM8K scores of different defense methods on the Llama3.1-8b model.}
\vspace{-0.7em}
\label{tab:model_comparison_llama3} 
\resizebox{0.9\textwidth}{!}{%
  \begin{tabular}{ 
      cc                    
    | *{5}{c} 
     c}
    \toprule
    \multirow{2}{*}{\textbf{Model}}
      & \multirow{2}{*}{\textbf{Defense}}
      & \multicolumn{5}{c}{\textbf{Just-Eval} (1–5)\,$\uparrow$} 
      & \multirow{2}{*}{\textbf{GSM8K} (\%)\,$\uparrow$} \\
      & 
      & \textbf{Helpfulness}
      & \textbf{Clarity}
      & \textbf{Factuality}
      & \textbf{Depth}
      & \textbf{Engaging}
      &  \\ 
    \midrule
    \multirow{3}{*}{Llama3.1-8b}
      & No Defense   & 4.23  & 4.89 & 4.80 & 3.97 & 4.35 & 81.8   \\
      & Deep-Align   & 3.44  & 4.50 & 3.77 & 3.32 & 4.06 & 10.3   \\
    \rowcolor{gray!8}
      & Ours          & 3.93  & 4.73 & 4.59 & 3.71 & 4.06 & 75.2   \\
    \bottomrule
  \end{tabular}%
}
\end{table*}